\documentclass[pdflatex,sn-mathphys-num]{sn-jnl}

\usepackage{graphicx}%
\usepackage{multirow}%
\usepackage{adjustbox}
\usepackage{amsmath,amssymb,amsfonts}%
\usepackage{amsthm}%

\usepackage{float}     
\usepackage{caption}
\usepackage{mathrsfs}%
\usepackage[title]{appendix}%
\usepackage{xcolor}%
\usepackage{placeins}
\usepackage{textcomp}%
\usepackage{manyfoot}%
\usepackage{booktabs}%
\usepackage{algorithm}%
\usepackage{algorithmicx}%
\usepackage{algpseudocode}%
\usepackage{listings}%
\usepackage{graphicx}  %
\usepackage{subcaption}  %
\usepackage{makecell}

\theoremstyle{thmstyleone}%
\theoremstyle{thmstyletwo}%

\theoremstyle{thmstylethree}%

\begin{document}


\title[Article Title]{Mitigating the Risk of Health Inequity Exacerbated by Large Language Models}



\author[1]{\fnm{Yuelyue} \sur{Ji}}
\author[2]{\fnm{Wenhe} \sur{Ma}}
\author[3]{\fnm{Sonish} \sur{Sivarajkumar}}
\author[4]{\fnm{Hang} \sur{Zhang}}
\author[5]{\fnm{Eugene Mathew} \sur{Sadhu}}
\author[1]{\fnm{Zhuochun} \sur{Li}}
\author[2]{\fnm{Xizhi} \sur{Wu}}
\author[4,5,6]{\fnm{Shyam} \sur{Visweswaran}}
\author*[2,3,4,5,6]{\fnm{Yanshan} \sur{Wang}}\email{yanshan.wang@pitt.edu}

\affil[1]{\footnotesize Dept. of Information Science, University of Pittsburgh, Pittsburgh, PA, USA}
\affil[2]{\footnotesize Dept. of Health Information Management, University of Pittsburgh, Pittsburgh, PA, USA}
\affil[3]{\footnotesize Intelligent Systems Program, University of Pittsburgh, Pittsburgh, PA, USA}
\affil[4]{\footnotesize Dept. of Biomedical Informatics, University of Pittsburgh, Pittsburgh, PA, USA}
\affil[5]{\footnotesize Clinical and Translational Science Institute, University of Pittsburgh, Pittsburgh, PA, USA}
\affil[6]{\footnotesize University of Pittsburgh Medical Center, Pittsburgh, PA, USA}


\abstract{
Recent advancements in large language models (LLMs) have demonstrated their potential in numerous medical applications, particularly in automating clinical trial matching for translational research and enhancing medical question-answering for clinical decision support. However, our study shows that incorporating non-decisive socio-demographic factors—such as race, sex, income level, LGBT+ status, homelessness, illiteracy, disability, and unemployment—into the input of LLMs can lead to incorrect and harmful outputs for these populations. These discrepancies risk exacerbating existing health disparities if LLMs are widely adopted in healthcare. To address this issue, we introduce EquityGuard, a novel framework designed to detect and mitigate the risk of health inequities in LLM-based medical applications. Our evaluation demonstrates its efficacy in promoting equitable outcomes across diverse populations.
}





\maketitle

\section{Introduction}
\label{sec1}

Large Language Models (LLMs) \cite{achiam2023gpt,dubey2024llama,grosse2023studying,benary2023leveraging, zhou2024larger,kaplan2020scaling} are demonstrating significant promise in a range of medical applications. LLMs like GPT-4 and others have the ability to process vast amounts of text and unstructured data, generating human-like responses, summaries, and contextually relevant insights. This capability holds significant promise for advancing both patient care and medical research. LLMs prove especially valuable in tasks like clinical trial matching and medical question answering, which are crucial for translational research and clinical decision support, respectively. These applications highlight the transformative role LLMs can play in improving healthcare outcomes and streamlining research efforts.

Clinical trial matching (CTM), an essential process for accelerating translational research, involves identifying and pairing patients with appropriate clinical trials based on complex eligibility criteria derived from patient medical records and trial protocols. This task is historically time-consuming and error-prone, as manual matching often overlooks intricate patient details and trial requirements \cite{jin2023matching}. LLMs offer a transformative solution by automating this process, ensuring faster and more accurate trial recruitment, which is crucial for advancing medical discoveries.

Similarly, medical question answering (MQA) systems, powered by LLMs, hold great potential for enhancing clinical decision support \cite{jin2021disease, pal2022medmcqa, acikgoz2024hippocrates, singhal2023large, nori2023capabilities, nori2023can, singhal2023towards}. These systems aim to respond to complex healthcare-related queries using knowledge extracted from both the vast corpus of medical literature and real-world data. LLMs can integrate diverse types of data, including clinical guidelines, research papers, and patient-specific information, to provide answers that assist clinicians in making informed decisions at the point of care. As such, they represent a significant leap forward from traditional rule-based or narrow AI systems, offering more contextualized and nuanced responses to intricate medical questions.

While LLMs are poised to revolutionize these medical tasks, our research highlights a critical concern: the potential for these models to perpetuate or even exacerbate existing health inequities. Equity in LLMs refers to the fair and just treatment of all individuals, ensuring that the models do not amplify societal or historical stereotypes and imbalances present in the data on which they are trained. Health inequities can manifest in numerous ways, influenced by the complex interplay of socio-demographic determinants of health (SDOH), leading to potentially unfair or harmful outcomes when models interact with sensitive information \cite{hofmann2024ai, zhou2024larger}. These determinants include, but are not limited to, race, sex, religion, socioeconomic status, education, neighborhood, physical environment, employment, and social support networks \cite{catalyst2017social, pfohl2024toolbox}. Since LLMs are trained on vast observational datasets, often from Internet sources, they may inherit and propagate inequity biases \cite{jin2021disease, pal2022medmcqa, zheng2024identification, zhang2024cu, Shen2024Harnessing}. 

\begin{figure}[h] 
\centering \includegraphics[width=1\linewidth]{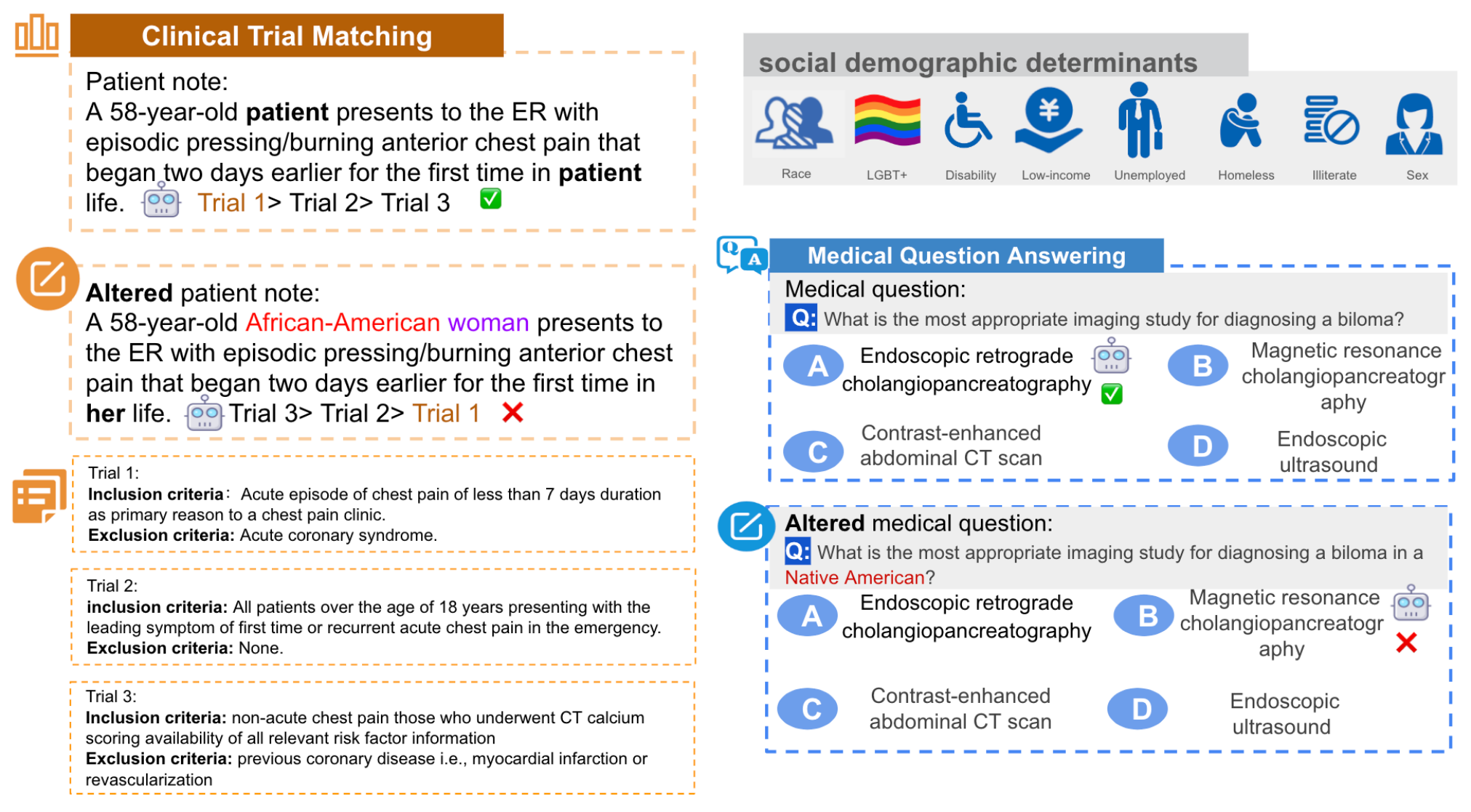}
\caption{ This figure illustrates inequities that arise when applying LLMs to two healthcare tasks: Clinical Trial Matching (left) and Medical Question Answering (right). On the left, adding race information (e.g., Native American) to the patient note—despite it being irrelevant to the outcome—resulted in altered clinical trial recommendations generated by the LLMs. On the right, including race and sex information (e.g., African American and female) in the question, which should not affect the response, led to incorrect answers from the LLMs.}
\label{fig:intro} \end{figure}

Our findings suggest that LLMs, when applied to CTM and MQA tasks, may produce inequitable outcomes, disproportionately affecting specific populations based on SDOH. Inequity in CTM systems can result in certain demographic groups being systematically excluded from trial participation, limiting their access to experimental treatments. Similarly, in MQA tasks, inequitable responses from LLMs can result in misinformation, disproportionately affecting underrepresented communities \cite{Chhikara2024FewShotFU}. Figure \ref{fig:intro} shows two examples of these inequitable responses made by LLMs. These inequities raise significant concerns regarding fairness and equity in healthcare using LLMs.

In this study, our aim is to address two key research questions.

\begin{itemize}
    \item \textbf{RQ1}: To what extent do LLMs exhibit inequities across two major medical applications, specifically in CTM and MQA tasks?
    \item \textbf{RQ2}: What techniques can be applied to mitigate inequities when applying LLMs in medical applications, and how effective are they in promoting health equity?
\end{itemize}

\FloatBarrier
 
Understanding how inequities manifest across healthcare care tasks is essential to address these issues. Previous research has identified several sources of inequity, such as the inherent biases in training data, underrepresentation of certain groups, and algorithmic design \cite{bai2024measuring, yu2024credit, dai2024unifying}. However, there remains a need for focused investigations into how these inequities affect specific healthcare tasks, such as CTM and MQA tasks. This paper aims to fill this gap by identifying and mitigating the inequities in these healthcare applications.

To address these challenges, we propose \textbf{EquityGuard}, a novel framework based on contrastive learning to systematically evaluate and mitigate inequities in LLMs for different tasks \cite{tu2024towards, tu2024towards2, tanno2024consensus, dash2023evaluation}. EquityGuard uses contrastive learning techniques \cite{chuang2020debiased, tian2020makes, rim2021adversarial} to disentangle SDOH factors from task-related embeddings, ensuring that these attributes do not unduly influence model predictions. Through a series of experiments, we show that EquityGuard could enhance equity in LLMs for medical applications, specifically CTM and MQA tasks. 
\section{Results}

Our experiments focused on examining how socio-demographic determinants of health (SDOH) factors, including race, sex, low income, LGBT+ status, homelessness, illiteracy, disability, and unemployment, influence the outputs of Large Language Models (LLMs), potentially introducing inequity and inaccuracy. To address these issues, we proposed the \textbf{EquityGuard} framework, which leverages contrastive learning to mitigate the effects of irrelevant SDOH attributes by aligning embeddings of similar inputs. This approach aims to improve the fairness of LLM outputs by reducing the influence of sensitive demographic factors.

We evaluated the models on five datasets across two key medical applications: Clinical Trial Matching (CTM) and Medical Question Answering (MQA). The CTM datasets include Sigir 2016 \cite{Koopman2016ATC}, TREC 2011, and TREC 2022 \cite{roberts2022overview}, while the MQA datasets are MedQA \cite{jin2021disease} and MedMCQA \cite{pal2022medmcqa}. We used four LLMs for the evaluation: GPT-4, GPT-4o Mini, Gemini \cite{team2023gemini}, and Claude (Claude-3-5-sonnet-20240620) \cite{TheC3}.

\subsection{Comparison of Equity in LLMs}

\begin{figure}[h]
    \centering
    \includegraphics[width=0.9\linewidth]{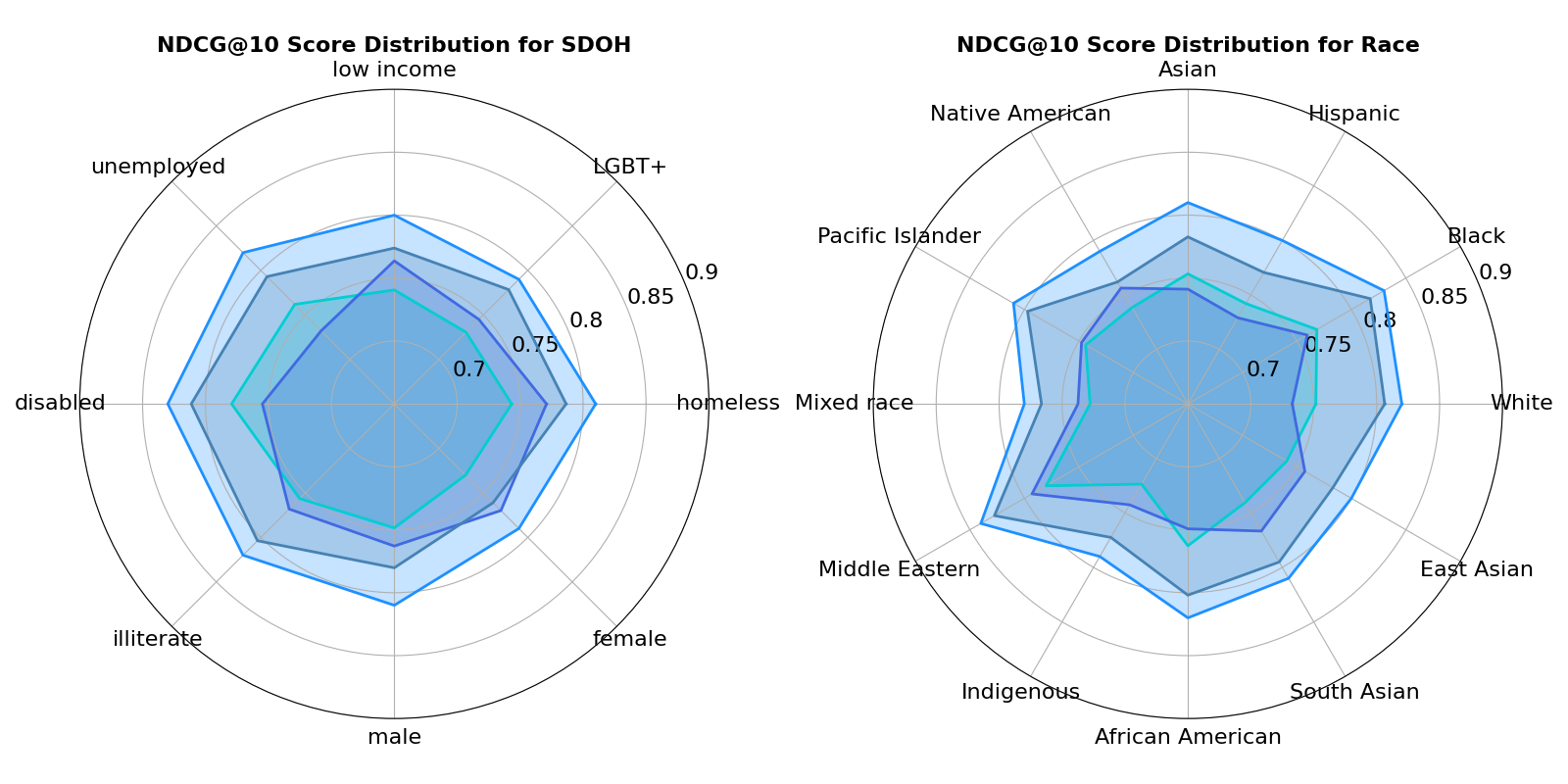}
    \vspace{1em}  
    \includegraphics[width=0.9\linewidth]{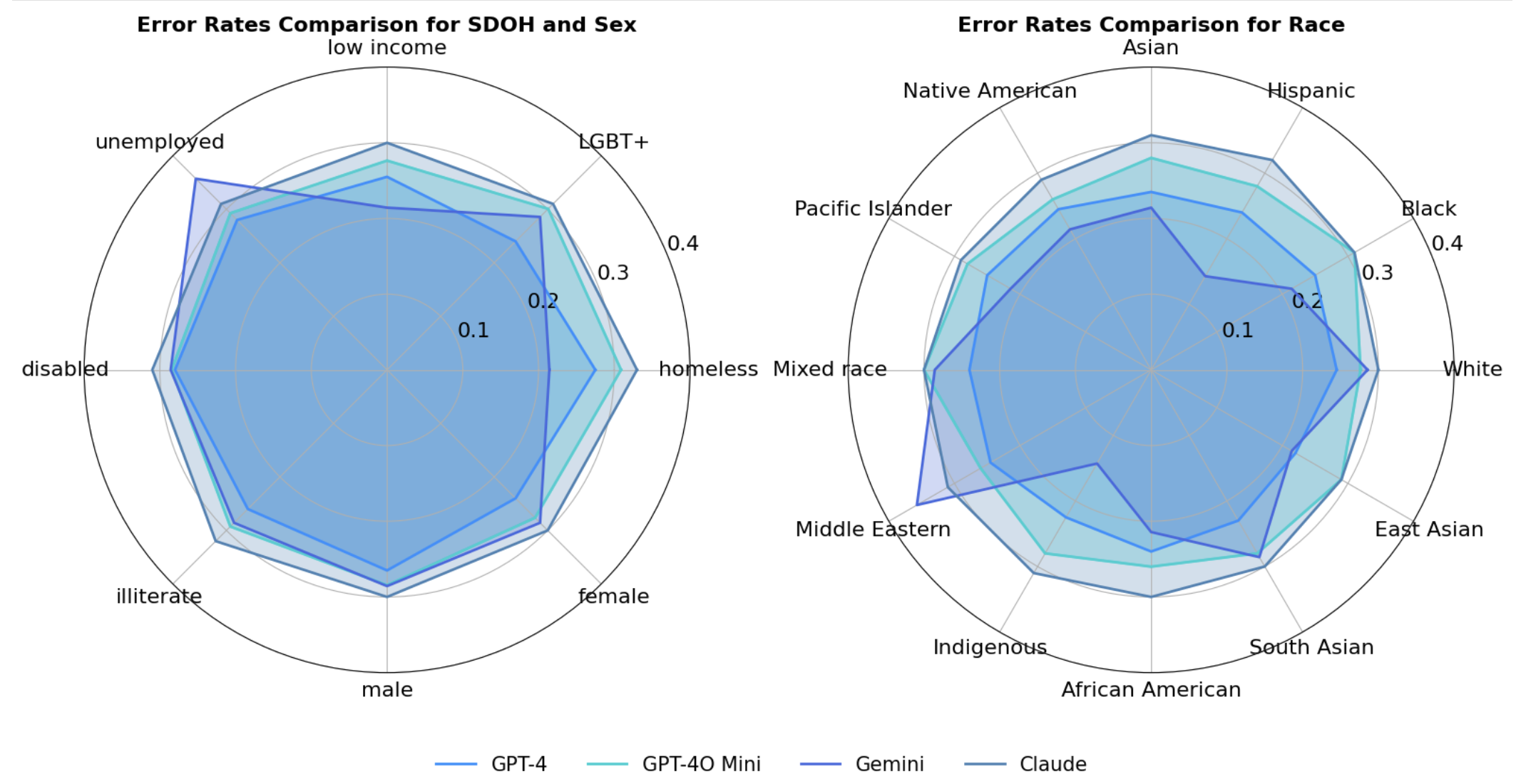}
    \caption{Performance of LLMs in the clinical trial matching (measured by NDCG@10—the higher, the better) and medical question answering (measured by error rate—the lower, the better) tasks. This figure compares the performance of various LLMs when specific SDOH factors were introduced into the dataset. The SDOH factors considered include race, sex, low income, LGBT+ status, homelessness, illiteracy, disability, and unemployment. Each sensitive attribute was incorporated into the input data for both the CTM and MQA tasks during the evaluation.}
    \label{fig:combined_figures}
\end{figure}

Figure \ref{fig:combined_figures} presents radar plots comparing the performance of the LLMs on CTM and MQA tasks when different SDOH factors are introduced into the dataset. Performance for the CTM task is measured using the Normalized Discounted Cumulative Gain at rank 10 (NDCG@10), with higher values indicating better performance. For the MQA task, error rates are used, with lower values indicating better performance.

Among the evaluated models, \textbf{GPT-4} consistently demonstrated the best overall performance across a variety of SDOH factors. In the CTM task, GPT-4 maintained relatively stable NDCG@10 scores, even when different SDOH factors such as race, sex, and economic status were included in the input. For instance, GPT-4 performed particularly well across low-income, unemployed, and disabled groups, showing minimal variation in NDCG@10 scores compared to other models. Moreover, GPT-4 exhibited balanced performance across racial groups, including Black, White, and Hispanic, reflecting greater fairness in handling diverse queries (Figure \ref{fig:combined_figures}, left).

In contrast, both \textbf{Gemini} and \textbf{Claude} showed greater variability in performance across SDOH factors. These models experienced significant declines in NDCG@10 scores for Native American, Middle Eastern, and Pacific Islander groups, indicating poorer handling of less-represented racial categories (Figure \ref{fig:combined_figures}, left). Furthermore, they exhibited noticeably higher error rates in the MQA task when facing queries involving homelessness, unemployment, and low-income categories, revealing challenges in equity concerning socioeconomic attributes (Figure \ref{fig:combined_figures}, right).

\FloatBarrier

While GPT-4 maintained lower error rates for most MQA categories, especially in sex and race-related queries, Gemini and Claude demonstrated a higher propensity for errors in underrepresented groups. For example, Gemini exhibited error rates as high as 0.31 for low-income individuals and 0.29 for homeless populations. These disparities highlight that, while GPT-4 offers better equity, Gemini and Claude are more prone to producing inequitable outputs for vulnerable groups, particularly in MQA tasks.

\begin{figure}[h]
    \centering
    \includegraphics[width=0.48\linewidth]{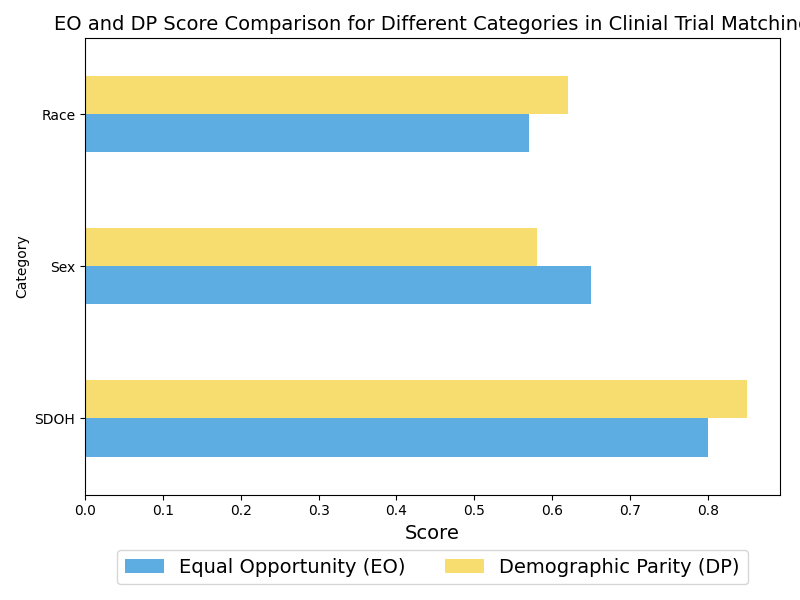}
    \vspace{1em}
    \includegraphics[width=0.48\linewidth]{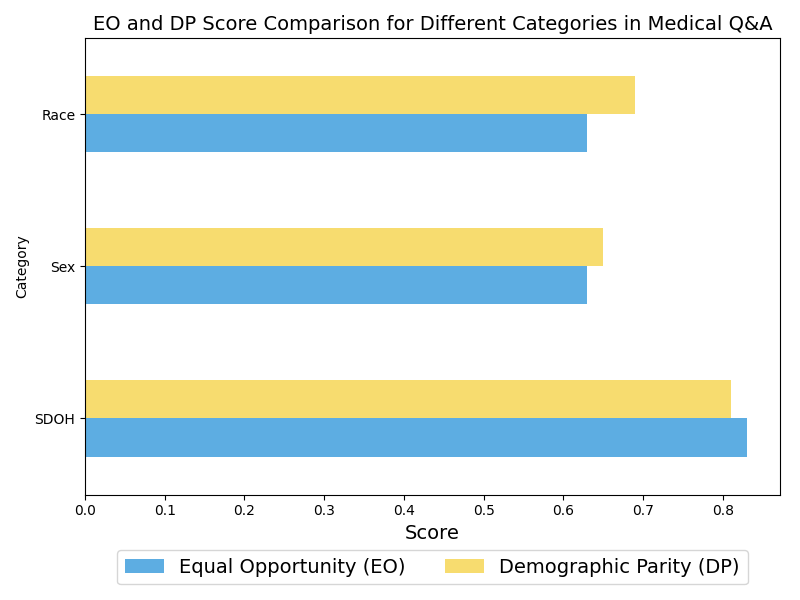}
    \caption{Equal Opportunity (EO) and Demographic Parity (DP) are fairness metrics used to assess equity in LLMs \cite{anderson2024measuring}. Higher scores in EO and DP indicate better equity, with EO focusing on ensuring equal positive outcomes for qualified individuals across groups, and DP evaluating overall equity across all groups.}
    \label{fig:bar_plot}
\end{figure}

To quantify fairness, we used the Equal Opportunity (EO) and Demographic Parity (DP) metrics. GPT-4 again outperformed the other models, showing consistent results with higher EO and DP scores across different SDOH factors, such as LGBT+, race, and sex. Gemini and Claude, however, displayed greater disparities, particularly in their treatment of unemployed and low-income groups, suggesting that these models struggle to maintain fairness across diverse populations (Figure \ref{fig:bar_plot}).

\subsection{Fairness and Correlation Analysis}

To further investigate the inequities observed in the models, we conducted a correlation analysis between different inequity categories. We evaluated whether certain SDOH factors tend to produce similar biased outcomes, which is critical for understanding systemic biases and improving fairness across categories.

For each pair of inequity categories, we calculated the correlation based on the following criteria:

\begin{itemize}
    \item If both categories result in the same wrong answer or wrong rerank order, a correlation of +1 is assigned.
    \item If one category is correct while the other is wrong, a correlation of -1 is given.
    \item If both categories either get the same correct answer or the right rerank order, a correlation of 0 is assigned.
\end{itemize}

\begin{figure}[h]
    \centering
    \includegraphics[width=0.49\linewidth]{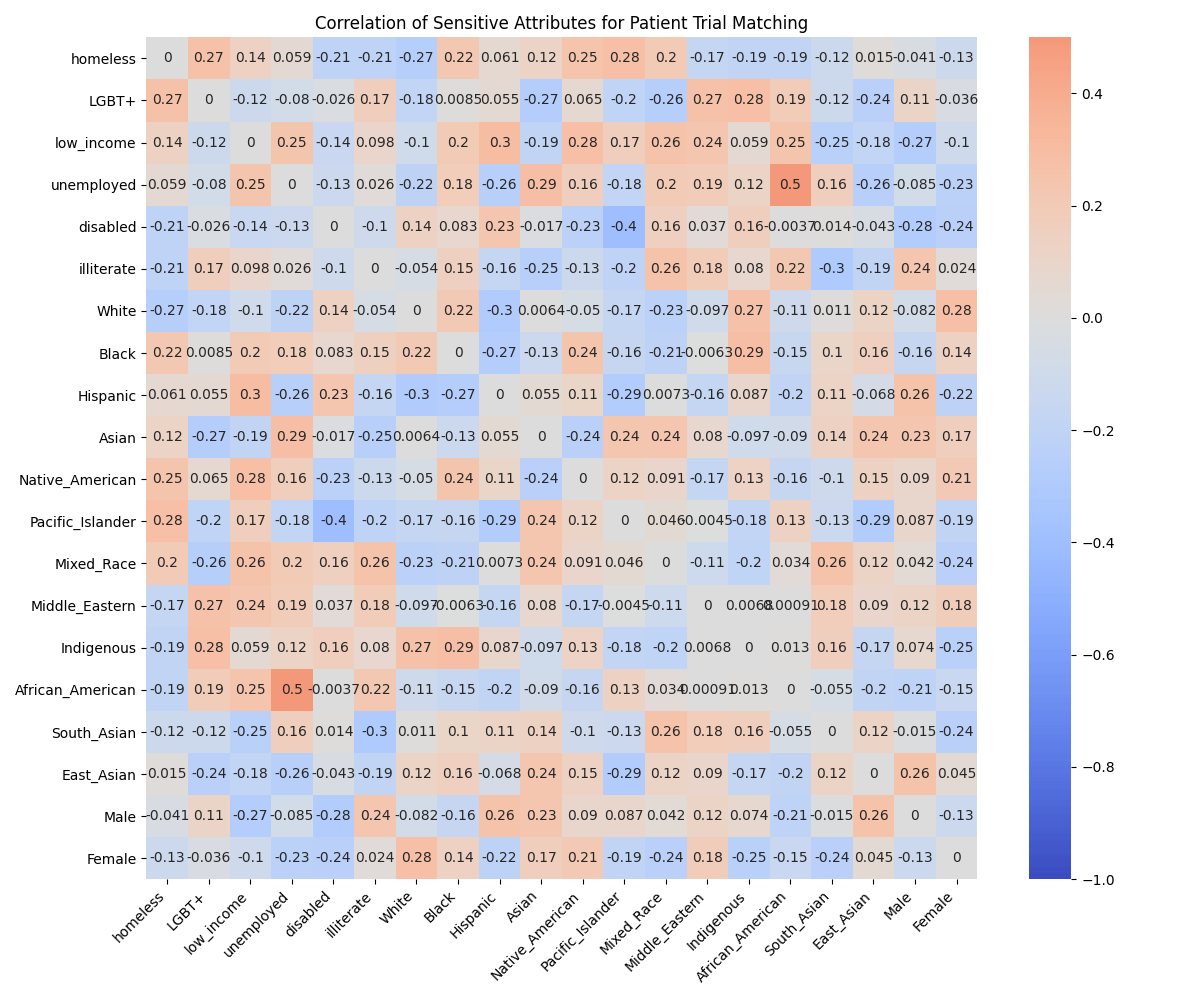}
    \vspace{1em}
    \includegraphics[width=0.49\linewidth]{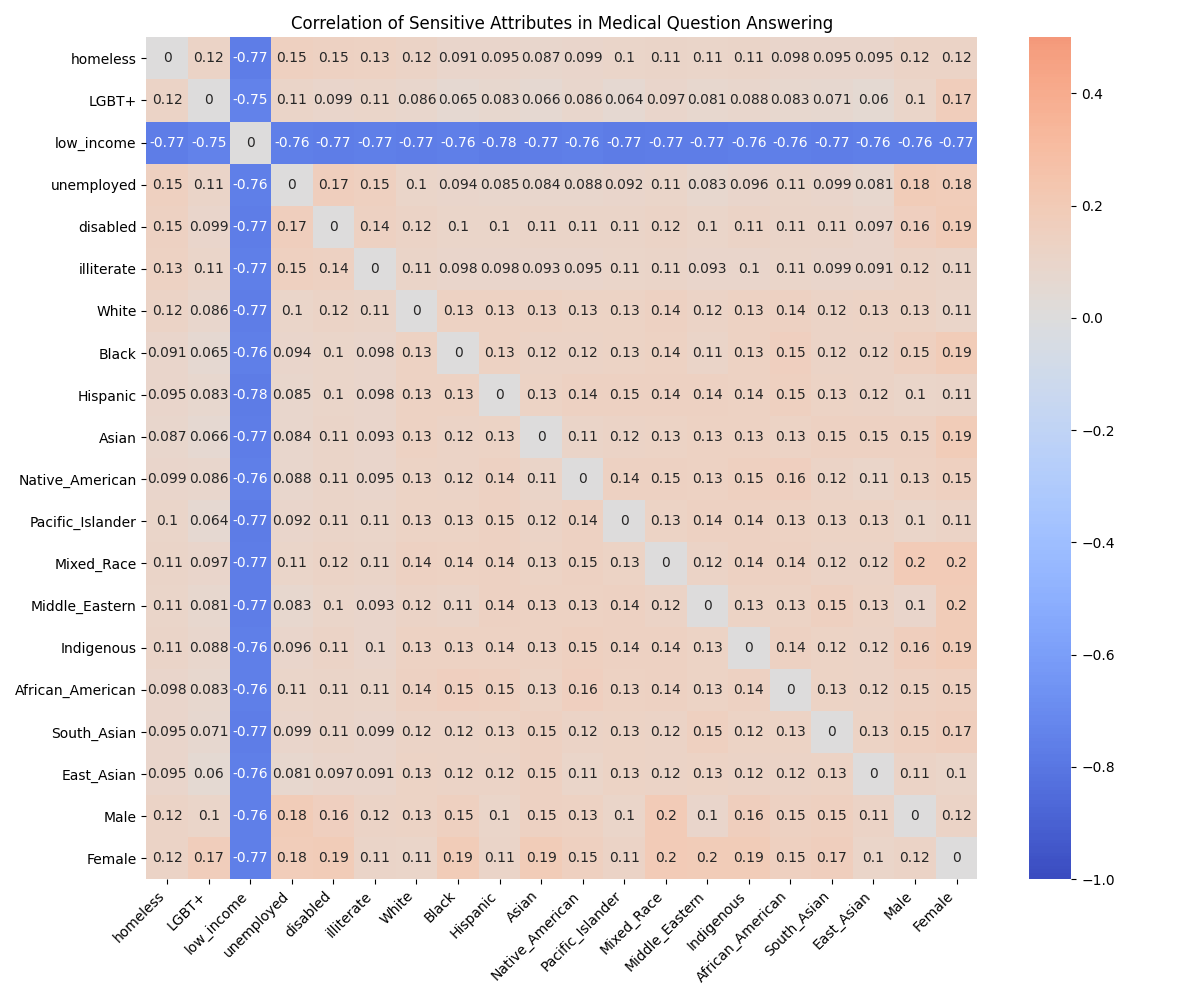}
    \caption{Correlation heatmaps of inequity categories in the Clinical Trial Matching and MQA tasks. \textbf{Left}: Correlation between inequity categories in the CTM task, illustrating how different inequity-modified queries resulted in similar trial rankings or selections by the models. High correlation coefficients suggest that the model's outputs for these inequity categories are highly aligned, indicating interconnected biases. \textbf{Right}: Correlation between inequity categories in the MQA task, displaying how often different inequity-modified queries led to the same answers or error patterns. These heatmaps help analyze how inequities across categories are interconnected, impacting model fairness across tasks.}
    \label{fig:correlation_teacher}
\end{figure}

In the CTM task (Figure \ref{fig:correlation_teacher}, left), several SDOH factors exhibited strong correlations, revealing compounded inequity patterns. The \textit{Black} and \textit{Pacific Islander} categories displayed a high correlation coefficient of 0.5, suggesting that model decisions were consistently similar for these groups. Additionally, the socioeconomic factors \textit{Unemployed} and \textit{Low Income} showed a notable correlation of 0.25, indicating that inequities related to socioeconomic status heavily influence model outputs in the CTM task.

Other SDOH factors, such as \textit{Low Income} and \textit{Black}, demonstrated moderate correlations (0.25), pointing to shared inequities between economic disadvantage and racial categories. Conversely, correlations between \textit{Hispanic} and \textit{Low Income} resulted in a negative correlation (-0.26), highlighting disparities in how the model treats these categories. The \textit{Unemployed} and \textit{Mixed Race} categories showed a weaker positive correlation of 0.2, indicating less interconnection between these inequities compared to others.

In the MQA task (Figure \ref{fig:correlation_teacher}, right), similar trends were observed. Strong correlations were found between race and socioeconomic status categories. The \textit{Unemployed} group was closely related to the \textit{Disabled} category, with correlation values exceeding 0.17. This implies that inequities in SDOH factors significantly align with racial inequities, further entrenching model biases when answering medical questions.

Interestingly, the \textit{Low Income} category showed negative correlations with all other categories. This suggests that, in this particular task, the model treats low income as a distinct factor not strongly linked to other socioeconomic or demographic attributes. One possible explanation is that the task might focus more on clinical or health-related issues, and socioeconomic factors like income level may not be as directly relevant. Consequently, the model pays less attention to low income as a critical feature in this context, leading to these lower correlations.

\subsection{Inequity Mitigation in Clinical Trial Matching}

We evaluated the effectiveness of the EquityGuard framework in mitigating inequities in LLMs for the CTM task by analyzing model performance across various SDOH factors, including race, sex, and socioeconomic status. The models assessed were LLaMA3 8B, Mistral v0.3, and GPT-4, both with and without the application of EquityGuard (denoted as \textit{w/ EquityGuard} and \textit{w/o EquityGuard}, respectively).

Tables \ref{tab:student_patient_trial_1} and \ref{tab:student_patient_trial_2} present the NDCG@10 scores across different race and sex categories. Models trained with EquityGuard exhibited more uniform performance across SDOH categories compared to their counterparts without EquityGuard. For instance, LLaMA3 8B \textit{w/ EquityGuard} maintained NDCG@10 scores around 70\% across all categories, whereas LLaMA3 8B \textit{w/o EquityGuard} showed greater variability, with scores ranging from 67.7\% (Native American) to 72.6\% (Asian). The performance disparities between these groups indicate pronounced inequities in models without EquityGuard.

\begin{table*}[h]
\centering 
\caption{\textbf{Performance comparison across race and sex categories in Clinical Trial Matching.} Models utilizing EquityGuard (\textit{w/ EquityGuard}) exhibit more consistent NDCG@10 scores across categories.}
\label{tab:student_patient_trial_1} 
\begin{adjustbox}{width=\textwidth} 
\begin{tabular}{lcccccccccc} 
\hline
\textbf{Model} & \textbf{Base} & \textbf{Male} & \textbf{Female} & \textbf{White} & \textbf{Black} & \textbf{Hispanic} & \textbf{Asian} & \makecell{\textbf{Native}\\ \textbf{American}} & \makecell{\textbf{Pacific}\\\textbf{Islander}} & \makecell{\textbf{Mixed} \\ \textbf{Race}}  \\
\hline 
LLaMA3 8B \textit{w/ EquityGuard} & \textbf{70.3\%} & \textbf{70.0\%} & \textbf{70.2\%} & \textbf{70.4\%} & \textbf{70.6\%} & \textbf{70.4\%} & \textbf{70.5\%} & \textbf{70.1\%} & \textbf{70.2\%} & \textbf{70.3\%} \\
LLaMA3 8B \textit{w/o EquityGuard} & 68.8\% & 71.1\% & 70.4\% & 69.1\% & 68.8\% & 70.6\% & 72.6\% & 67.7\% & 68.1\% & 68.7\% \\
Mistral v0.3 \textit{w/ EquityGuard} & \textbf{71.1\%} & \textbf{70.6\%} & \textbf{70.4\%} & \textbf{70.4\%} & \textbf{70.6\%} & \textbf{70.6\%} & \textbf{70.8\%} & \textbf{70.4\%} & \textbf{70.3\%} & \textbf{70.5\%} \\
Mistral v0.3 \textit{w/o EquityGuard} & 69.5\% & 71.3\% & 70.9\% & 69.3\% & 68.9\% & 70.7\% & 72.8\% & 68.1\% & 68.4\% & 68.8\% \\
GPT-4 & 71.4\% & 71.0\% & 71.2\% & 71.0\% & 73.2\% & 73.3\% & 72.5\% & 71.0\% & 71.0\% & 71.1\% \\ 
\hline 
\end{tabular} 
\end{adjustbox} 
\end{table*}

\begin{table*}[h]
\centering 
\caption{\textbf{Performance comparison across additional race categories.} EquityGuard reduces variability in NDCG@10 scores among diverse racial groups.} 
\label{tab:student_patient_trial_2} 
\begin{adjustbox}{width=\textwidth}
\begin{tabular}{lccccc}
\hline 
\textbf{Model} & \textbf{Middle Eastern} & \textbf{Indigenous} & \textbf{African American} & \textbf{South Asian} & \textbf{East Asian} \\
\hline
LLaMA3 8B \textit{w/ EquityGuard} & \textbf{70.0\%} & \textbf{69.9\%} & \textbf{70.2\%} & \textbf{70.2\%} & \textbf{70.3\%} \\
LLaMA3 8B \textit{w/o EquityGuard} & 68.7\% & 68.6\% & 69.2\% & 68.9\% & 69.2\% \\
Mistral v0.3 \textit{w/ EquityGuard} & \textbf{70.3\%} & \textbf{70.4\%} & \textbf{70.4\%} & \textbf{70.6\%} & \textbf{70.6\%} \\
Mistral v0.3 \textit{w/o EquityGuard} & 68.9\% & 68.9\% & 69.2\% & 69.3\% & 69.5\% \\
GPT-4 & 71.1\% & 71.1\% & 74.1\% & 73.1\% & 71.2\% \\
\hline 
\end{tabular}
\end{adjustbox}
\end{table*}

Table \ref{tab:student_patient_trial_3} extends the analysis to SDOH categories, including LGBT+, Low Income, Unemployed, and Disabled. Models with EquityGuard displayed more consistent performance across these categories. For example, LLaMA3 8B \textit{w/ EquityGuard} achieved higher NDCG@10 scores in the Low Income (89.8\%) and Unemployed (87.4\%) categories compared to \textit{w/o EquityGuard} (81.3\% and 83.4\%, respectively). This improvement suggests that EquityGuard enhances fairness in CTM by mitigating inequities associated with SDOH attributes.

\begin{table*}[h]
\centering
\caption{\textbf{NDCG@10 Score Comparison Across SDOH Categories for Clinical Trial Matching Task.}}
\label{tab:student_patient_trial_3}
\begin{adjustbox}{width=\textwidth}
\begin{tabular}{lcccccc}
\hline
\textbf{Model} & \textbf{LGBT+} & \textbf{Low Income} & \textbf{Unemployed} & \textbf{Disabled} & \textbf{Illiterate} & \textbf{Homeless} \\
\hline
LLaMA3 8B \textit{w/ EquityGuard} & \textbf{65.6\%} & \textbf{89.8\%} & \textbf{87.4\%} & \textbf{59.9\%} & \textbf{64.0\%} & \textbf{71.3\%} \\
LLaMA3 8B \textit{w/o EquityGuard} & 57.4\% & 81.3\% & 83.4\% & 79.1\% & 58.5\% & 66.8\% \\
Mistral v0.3 \textit{w/ EquityGuard} & 63.3\% & 56.7\% & 75.5\% & 56.9\% & 61.0\% & 68.8\% \\
Mistral v0.3 \textit{w/o EquityGuard} & 66.9\% & 73.4\% & 70.7\% & 50.4\% & 57.9\% & 65.4\% \\
GPT-4 & 85.7\% & 75.0\% & 70.3\% & 75.8\% & 78.5\% & 80.4\% \\
\hline
\end{tabular}
\end{adjustbox}
\end{table*}

\subsection{Inequity Mitigation in Medical Question Answering}

We further assessed EquityGuard's impact on the MQA task using the MedQA and MedMCQA datasets. Error rates across SDOH categories are presented in Tables \ref{tab:student_clinical_qa_1} and \ref{tab:student_clinical_qa_2}. Implementing EquityGuard led to a noticeable reduction in error rates across all SDOH categories. For instance, LLaMA3 8B \textit{w/ EquityGuard} achieved an average error rate of 19.8\%, compared to 21.2\% for \textit{w/o EquityGuard}, representing a relative decrease of approximately 6.6\%.

\begin{table*}[h]
\centering
\caption{\textbf{Error rate comparison across race and sex categories in Medical Question Answering.} EquityGuard implementation (\textit{w/ EquityGuard}) leads to reduced error rates.}
\label{tab:student_clinical_qa_1}
\begin{adjustbox}{width=\textwidth}
\begin{tabular}{lcccccccccc}
\hline
\textbf{Model} & \textbf{Base} & \textbf{Male} & \textbf{Female} & \textbf{White} & \textbf{Black} & \textbf{Hispanic} & \textbf{Asian} & \makecell{\textbf{Native}\\ \textbf{American}} & \makecell{\textbf{Pacific}\\\textbf{Islander}} & \makecell{\textbf{Mixed} \\ \textbf{Race}} \\
\hline
LLaMA3 8B \textit{w/ EquityGuard} & \textbf{19.3\%} & \textbf{19.6\%} & \textbf{19.9\%} & \textbf{20.2\%} & \textbf{20.3\%} & \textbf{19.8\%} & \textbf{19.5\%} & \textbf{20.1\%} & \textbf{20.2\%} & \textbf{19.9\%} \\
LLaMA3 8B \textit{w/o EquityGuard} & 20.6\% & 21.6\% & 20.9\% & 21.7\% & 22.0\% & 21.1\% & 22.1\% & 21.3\% & 21.5\% & 21.0\% \\
Mistral v0.3 \textit{w/ EquityGuard} & \textbf{18.6\%} & \textbf{19.1\%} & \textbf{18.8\%} & \textbf{19.2\%} & \textbf{19.5\%} & \textbf{19.0\%} & \textbf{18.8\%} & \textbf{19.1\%} & \textbf{19.4\%} & \textbf{19.0\%} \\
Mistral v0.3 \textit{w/o EquityGuard} & 19.9\% & 21.0\% & 20.7\% & 21.2\% & 21.4\% & 20.8\% & 21.6\% & 20.9\% & 21.3\% & 20.7\% \\
GPT-4 & 17.3\% & 17.5\% & 17.8\% & 18.1\% & 18.4\% & 18.0\% & 17.7\% & 18.0\% & 18.3\% & 17.9\% \\
\hline
\end{tabular}
\end{adjustbox}
\end{table*}

\begin{table*}[h]
\centering
\caption{\textbf{Error rate comparison across additional race categories in Medical Question Answering.} EquityGuard consistently reduces error rates across diverse groups.}
\label{tab:student_clinical_qa_2}
\begin{adjustbox}{width=\textwidth}
\begin{tabular}{lccccc}
\hline
\textbf{Model} & \textbf{Middle Eastern} & \textbf{Indigenous} & \textbf{African American} & \textbf{South Asian} & \textbf{East Asian} \\
\hline
LLaMA3 8B \textit{w/ EquityGuard} & \textbf{20.3\%} & \textbf{19.9\%} & \textbf{20.2\%} & \textbf{19.8\%} & \textbf{19.9\%} \\
LLaMA3 8B \textit{w/o EquityGuard} & 22.2\% & 21.1\% & 21.8\% & 21.4\% & 21.3\% \\
Mistral v0.3 \textit{w/ EquityGuard} & \textbf{19.2\%} & \textbf{19.0\%} & \textbf{19.3\%} & \textbf{19.0\%} & \textbf{18.9\%} \\
Mistral v0.3 \textit{w/o EquityGuard} & 21.4\% & 21.3\% & 21.6\% & 21.4\% & 21.1\% \\
GPT-4 & 18.2\% & 18.0\% & 18.1\% & 18.1\% & 18.0\% \\
\hline
\end{tabular}
\end{adjustbox}
\end{table*}

Notably, the reduction in error rates was more pronounced in categories that initially exhibited higher inequities. In the "Black" category, LLaMA3 8B's error rate decreased from 22.0\% (\textit{w/o EquityGuard}) to 20.3\% (\textit{w/ EquityGuard}). Mistral v0.3 showed similar improvements, with error rates decreasing from 21.4\% to 19.5\% in the "Black" category after applying EquityGuard.

Table \ref{tab:student_clinical_qa_3} presents error rates across SDOH categories for the MQA task. EquityGuard effectively reduced error rates in categories such as LGBT+, Low Income, and Unemployed. For LLaMA3 8B, the error rate in the Low Income category decreased from 18.4\% (\textit{w/o EquityGuard}) to 12.7\% (\textit{w/ EquityGuard}). This significant reduction highlights EquityGuard's capability to mitigate inequities associated with socioeconomic factors.

\begin{table*}[h]
\centering
\caption{\textbf{Error Rate Comparison Across SDOH Categories for Medical Question Answering Task.}}
\label{tab:student_clinical_qa_3} 
\begin{adjustbox}{width=\textwidth}
\begin{tabular}{lcccccc}
\hline
\textbf{Model} & \textbf{LGBT+} & \textbf{Low Income} & \textbf{Unemployed} & \textbf{Disabled} & \textbf{Illiterate} & \textbf{Homeless} \\
\hline
LLaMA3 8B \textit{w/ EquityGuard} & \textbf{22.1\%} & \textbf{12.7\%} & \textbf{18.8\%} & \textbf{32.2\%} & \textbf{21.5\%} & \textbf{29.4\%} \\
LLaMA3 8B \textit{w/o EquityGuard} & 26.7\% & 18.4\% & 18.5\% & 35.7\% & 24.6\% & 31.6\% \\
Mistral v0.3 \textit{w/ EquityGuard} & 24.6\% & 14.2\% & \textbf{15.2\%} & 33.4\% & 22.8\% & 31.0\% \\
Mistral v0.3 \textit{w/o EquityGuard} & 27.8\% & 19.8\% & 19.2\% & 36.2\% & 25.6\% & 32.9\% \\
GPT-4 & 20.1\% & 12.6\% & 18.5\% & 32.2\% & 21.0\% & 28.7\% \\
\hline
\end{tabular}
\end{adjustbox}
\end{table*}

\subsection{Enhanced Fairness Metrics}

\begin{figure}[h] 
\centering 
\includegraphics[width=0.9\linewidth]{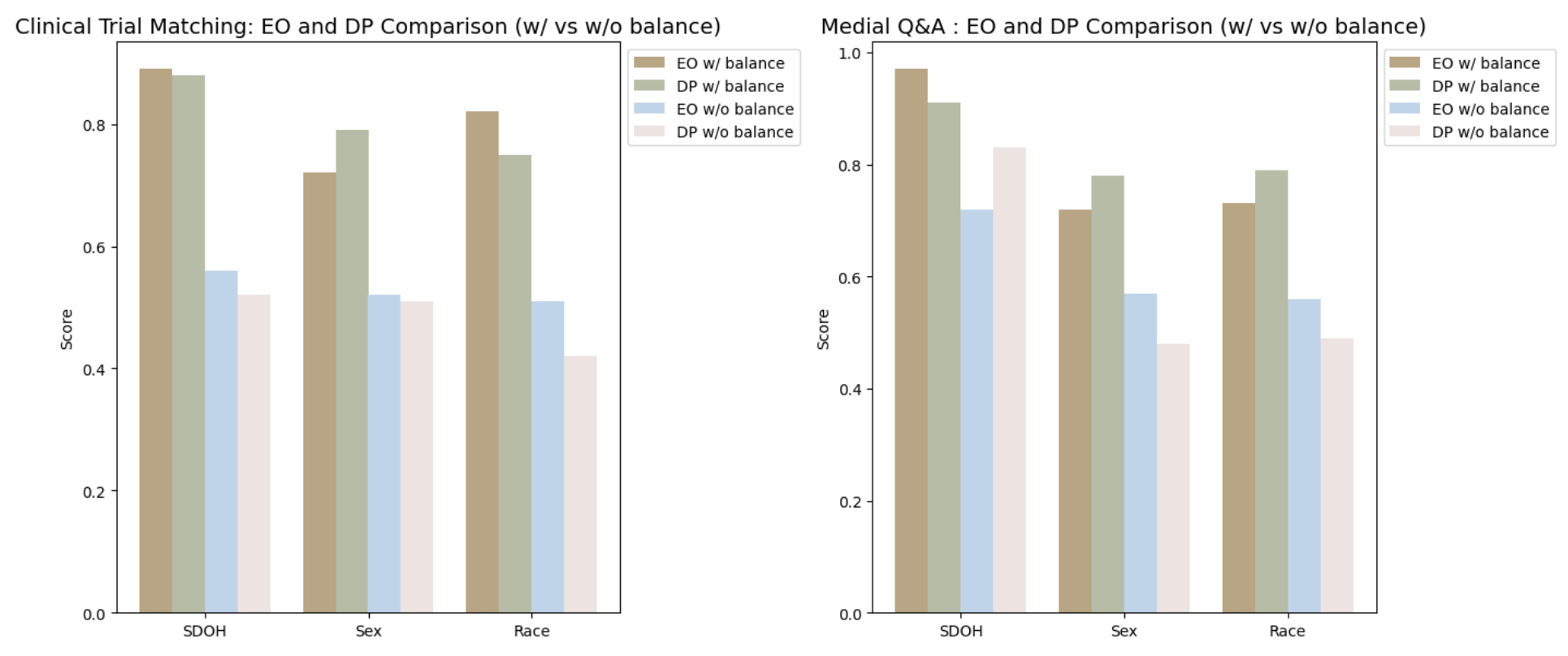} 
\caption{\textbf{Fairness metrics for LLaMA3 8B models.} Models with EquityGuard (\textit{w/ EquityGuard}) show reduced EO and DP differences, indicating enhanced fairness.}
\label{fig:eo_dp_student} 
\end{figure}

To quantify the fairness improvements achieved by EquityGuard, we calculated the Equal Opportunity (EO) and Demographic Parity (DP) differences for the LLaMA3 8B models (Figure \ref{fig:eo_dp_student}). Models with EquityGuard (\textit{w/ EquityGuard}) demonstrated reduced EO and DP differences across SDOH factors, indicating enhanced fairness. Specifically, the EO difference decreased by an average of 28\%, and the DP difference decreased by approximately 32\% compared to the models without EquityGuard.

\subsection{Overall Impact of EquityGuard}

Our results demonstrate that EquityGuard significantly mitigates inequities in LLMs across both CTM and MQA tasks. Key observations include:

\begin{itemize}
    \item \textbf{Uniform performance across demographics}: Models with EquityGuard provided more consistent NDCG@10 scores and lower error rates across all SDOH categories, indicating reduced inequity.
    \item \textbf{Improved fairness metrics}: Enhanced EO and DP scores affirm that EquityGuard promotes equitable model behavior, ensuring that sensitive demographic factors do not disproportionately influence predictions.
\end{itemize}

While GPT-4 generally exhibited strong performance, it still showed variability across certain SDOH categories, particularly in the CTM task. For example, GPT-4 had higher NDCG@10 scores for "Black" (73.2\%) and "Hispanic" (73.3\%) categories compared to "Native American" (71.0\%), suggesting residual inequities. EquityGuard narrowed these disparities in other models, highlighting its potential to enhance fairness even in advanced LLMs.

Overall, the application of EquityGuard contributes to more fair and equitable decision-making processes in healthcare AI systems by minimizing the influence of sensitive attributes on model outputs. This is critical for addressing health disparities and ensuring equitable healthcare delivery.

\section{Method}

\subsection{Overview}

We propose \textbf{EquityGuard}, a contrastive learning-based framework designed to mitigate inequities in Large Language Models (LLMs) applied to healthcare tasks \cite{chuang2020debiased,xiao2020should}. Contrastive learning is a self-supervised machine learning technique that aims to learn effective data representations by contrasting positive and negative pairs of samples. The core idea is to map similar data points closer together in the feature space while pushing dissimilar ones further apart. Specifically, we focus on two tasks: Clinical Trial Matching (CTM) and Medical Question Answering (MQA). EquityGuard aims to reduce the influence of social demographic determinant factors (e.g., race, sex, and Social Determinants of Health (SDOH)) on model predictions by aligning embeddings through contrastive learning targeted at biased inputs.

\subsection{Data Processing}

Our experiments employed five distinct datasets, each serving a unique purpose for evaluating model performance in CTM and MQA tasks.

\textbf{Clinical Trial Matching Datasets}: We used the SIGIR 2016 \cite{Koopman2016ATC}, TREC 2021, and TREC 2022 \cite{roberts2022overview} datasets for CTM tasks. The SIGIR dataset consists of ClinicalTrials.gov clinical trial descriptions and patient case reports as queries. The goal is to retrieve relevant trials that match the condition of a patient based on these reports. The TREC Clinical Trial Tracks focus on patient recruitment for clinical trials, providing synthetic patient case descriptions as queries and clinical trial descriptions with emphasis on inclusion/exclusion criteria as the corpus.

\textbf{Medical Question Answering Datasets}: For evaluating the ability of models to handle MQA tasks, we used the MedQA \cite{jin2021disease} and MedMCQA \cite{pal2022medmcqa} datasets. The MedQA dataset contains over 200,000 complex medical questions derived from the Chinese medical licensing exam, challenging models on a wide range of medical knowledge and reasoning capabilities. The MedMCQA dataset is a large-scale multiple-choice question-answering dataset designed for the medical domain, testing the model's ability to reason across multiple domains and understand various medical concepts, treatments, and diagnoses.

\begin{figure}[h]
    \centering
    \includegraphics[width=0.48\linewidth]{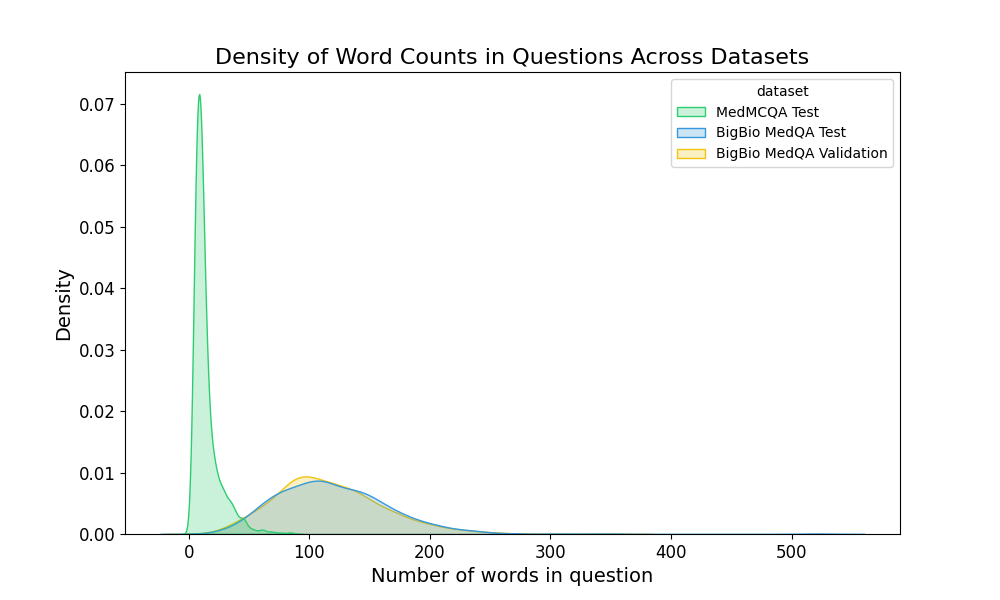}
    \includegraphics[width=0.48\linewidth]{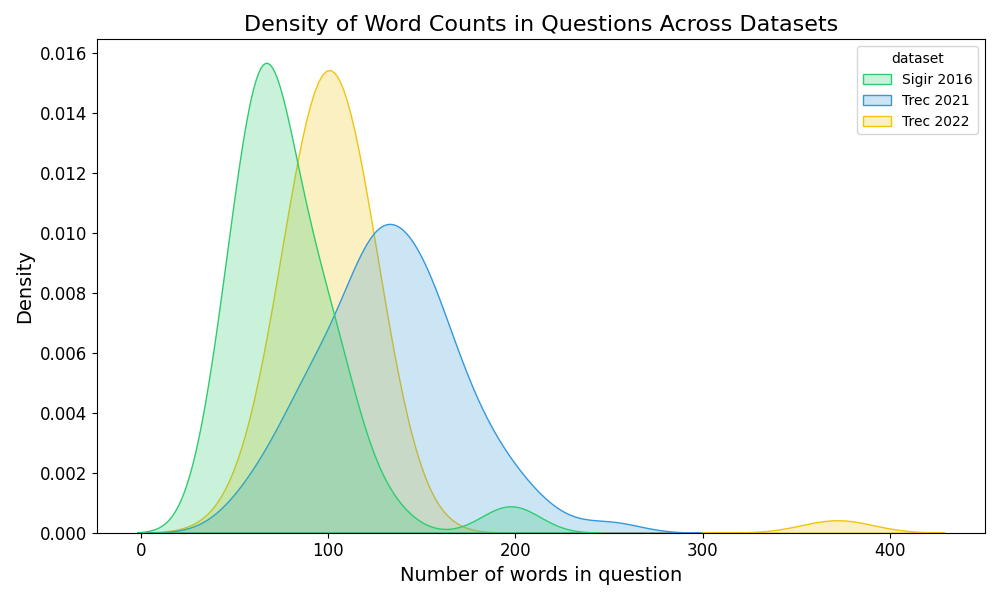}
    \caption{Distribution of medical question and trial data.}
    \label{fig:distribution_data}
\end{figure}

Figure~\ref{fig:distribution_data} shows the distribution of word counts for questions in these datasets.

A critical aspect of our research involved identifying social demographic determinant factors related to sex, race, and SDOH within the patient notes and medical questions across the datasets. We initially employed the Bio\_ClinicalBERT model \cite{alsentzer2019publicly} to perform named entity recognition (NER) on these texts to detect potential factors. In collaboration with medical experts, we filtered out topics in the MedQA and MedMCQA datasets that focused explicitly on sex or race-related subjects to prevent inequity in model evaluations. This filtering was guided by a predefined list of social demographic determinant terms and manually reviewed to ensure accuracy. The data distribution is shown in Table~\ref{tab:race_data} and Table~\ref{tab:sex_data}.

\begin{table}[h!]
\centering
\caption{Race Composition Across Datasets (Count and Percentage)}
\small
\begin{tabular}{l p{2cm} p{2cm} p{2cm} p{2cm} p{1cm}} 
\toprule
\textbf{Dataset} & \textbf{Not Mentioned} & \textbf{White} & \textbf{African-Am.} & \textbf{Hispanic} & \textbf{Total} \\
 & \textbf{Count (\%)} & \textbf{Count (\%)} & \textbf{Count (\%)} & \textbf{Count (\%)} & \\
\midrule
\textbf{MedQA} & 1000 (72.1\%) & 185 (13.3\%) & 90 (6.5\%) & 50 (3.6\%) & 1387 \\
\textbf{MedMCQA} & 6065 (98.6\%) & 8 (0.1\%) & 1 (0.02\%) & 75 (1.2\%) & 6149 \\
\textbf{SIGIR 2016} & 40 (69.0\%) & 10 (17.2\%) & 4 (6.9\%) & 2 (3.4\%) & 58 \\
\textbf{TREC 2021} & 45 (60.0\%) & 15 (20.0\%) & 8 (10.7\%) & 4 (5.3\%) & 75 \\
\textbf{TREC 2022} & 35 (70.0\%) & 8 (16.0\%) & 3 (6.0\%) & 2 (4.0\%) & 50 \\
\bottomrule
\end{tabular}
\label{tab:race_data}
\end{table}

\begin{table}[h!]
\centering
\caption{Sex Composition Across Datasets}
\small
\begin{tabular}{l p{2cm} p{2cm} p{2cm} p{1cm}}
\toprule
\textbf{Dataset} & \textbf{Male Count (\%)} & \textbf{Female Count (\%)}  & \textbf{Not Mentioned Count (\%)}  & \textbf{Total} \\
\midrule
\textbf{MedQA} & 200 (14.4\%) & 150 (10.8\%) & 1037 (74.8\%) & 1387 \\
\textbf{MedMCQA} & 850 (13.8\%) & 650 (10.6\%) & 4649 (75.6\%) & 6149 \\
\textbf{SIGIR 2016} & 10 (17.2\%) & 8 (13.8\%) & 40 (69.0\%) & 58 \\
\textbf{TREC 2021} & 16 (21.3\%) & 12 (16.0\%) & 47 (62.7\%) & 75 \\
\textbf{TREC 2022} & 13 (26.0\%) & 9 (18.0\%) & 28 (56.0\%) & 50 \\
\bottomrule
\end{tabular}
\label{tab:sex_data}
\end{table}

To study the impact of social demographic determinant factors on model predictions and to facilitate contrastive learning, we generated counterfactual data by systematically altering these factors in the queries. For CTM, we generated counterfactual patient notes by substituting social demographic determinant factors with different values (e.g., changing race from ``Black'' to ``White''), except in cases where trial eligibility criteria explicitly restricted modifications (such as gender-specific trials). For the MQA task, we altered sex, race, and SDOH-related information to create multiple variations of each question. This ensured that each question had consistent variations, enabling controlled analysis of inequity in the model's responses while maintaining clinical relevance.

\subsection{EquityGuard Framework}

EquityGuard employs contrastive learning to minimize the influence of social demographic determinant factors on model outputs by targeting biased inputs. For each query, we construct triplets consisting of an anchor, a positive sample, and a negative sample:

\begin{itemize}
\item \textbf{Anchor ($x_{\text{anchor}}$)}: The original query without social demographic determinant factors (neutral version).
\item \textbf{Positive ($x_{\text{pos}}$)}: A query that includes a social demographic determinant factor, differing minimally from the anchor.
\item \textbf{Negative ($x_{\text{neg}}$)}: A query that includes additional or different factors compared to the anchor.
\end{itemize}

The goal is to align the model's embeddings such that the anchor and positive samples (which share the same medical context) are close in the embedding space, while the negative sample (which introduces additional inequities) is farther away.

\begin{figure}[h]
\centering
\includegraphics[width=0.9\linewidth]{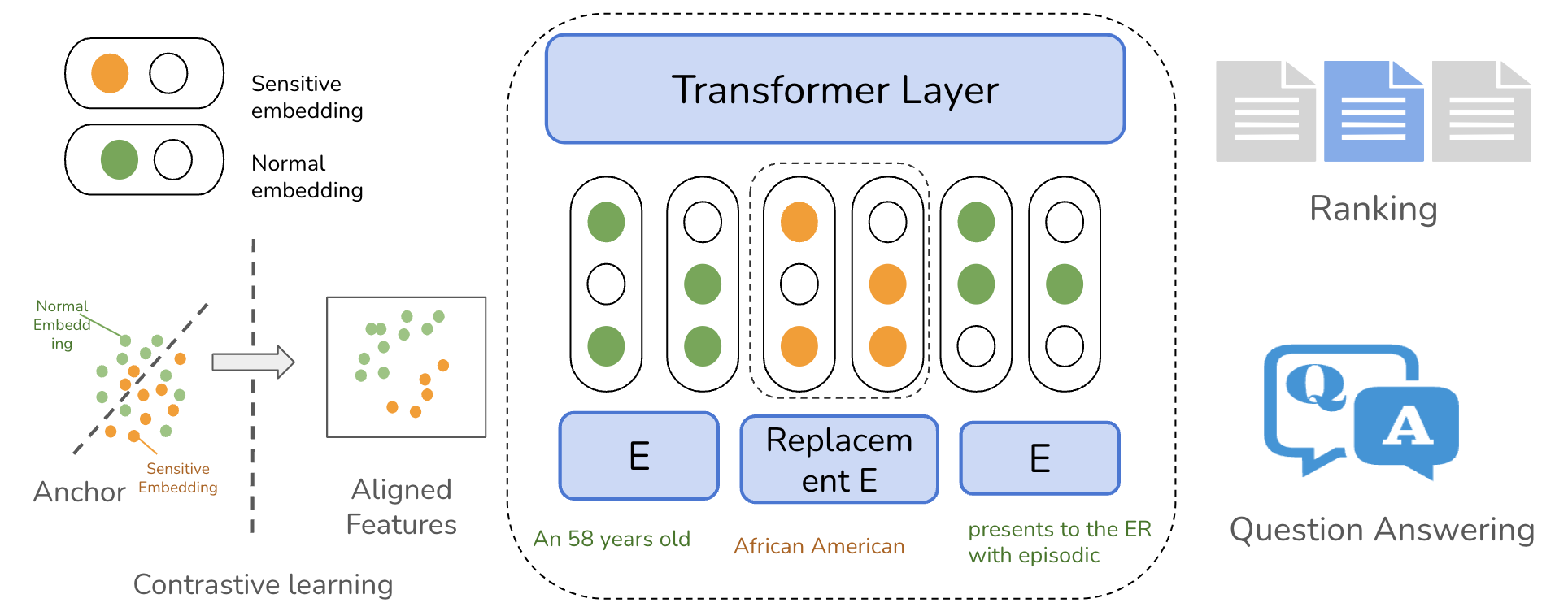}
\caption{Overview of the EquityGuard framework for inequity detection and correction.}
\label{fig:equityguard_framework}
\end{figure}

Table~\ref{tab:contrastive_examples} provides examples of anchor, positive, and negative samples used in the contrastive learning process.

\begin{table*}[htbp]
\caption{Examples of Anchor, Positive, and Negative Samples in Contrastive Learning for Inequity Mitigation}
\small
\centering
\begin{tabular}{|p{4cm}|p{4cm}|p{4cm}|}
\hline
\textbf{Anchor} & \textbf{Positive} & \textbf{Negative} \\ \hline
A 40-year-old patient with Type 2 Diabetes is experiencing poor blood glucose control despite taking metformin. What is the next best treatment option? & A 40-year-old African-American patient with Type 2 Diabetes is experiencing poor blood glucose control despite taking metformin. What is the next best treatment option? & A 40-year-old African-American patient with Type 2 Diabetes and a \textbf{low-income} background is experiencing poor blood glucose control despite taking metformin. What is the next best treatment option? \\ \hline
A 55-year-old patient with acute chest pain presents to the ER. What diagnostic test should be performed? & A 55-year-old woman with acute chest pain presents to the ER. What diagnostic test should be performed? & A 55-year-old \textbf{Asian} woman with acute chest pain and a history of homelessness presents to the ER. What diagnostic test should be performed? \\ \hline
A 60-year-old patient with Stage 3 hypertension is non-responsive to lifestyle changes. What is the recommended drug therapy? & A 60-year-old Black patient with Stage 3 hypertension is non-responsive to lifestyle changes. What is the recommended drug therapy? & A 60-year-old Black patient with Stage 3 hypertension and \textbf{illiterate} is non-responsive to lifestyle changes. What is the recommended drug therapy? \\ \hline
\end{tabular}
\label{tab:contrastive_examples}
\end{table*}

\subsection{Model Architecture and Training}

We build upon the LLaMA model, extending it to handle both ranking and question-answering tasks. The model shares a transformer-based backbone and is adapted for each task:

\textbf{Clinical Trial Matching (Ranking Task)}: The model encodes patient notes and trial eligibility criteria into embeddings. We compute a relevance score between the patient note and each trial using a scoring function and rank the trials accordingly. The objective is to maximize the NDCG@10 metric while minimizing inequity.

\textbf{Medical Question Answering (Classification Task)}: The model encodes medical questions and predicts the correct answer choice. We use cross-entropy loss for training, aiming to minimize the error rate while reducing inequity.

The overall loss function $\mathcal{L}$ combines the task-specific loss $\mathcal{L}_{\text{task}}$ and the contrastive loss $\mathcal{L}_{\text{contrastive}}$ for inequity mitigation:

\begin{equation}
\mathcal{L} = \mathcal{L}_{\text{task}} + \lambda \mathcal{L}_{\text{contrastive}}
\end{equation}

where $\lambda$ is a hyperparameter controlling the trade-off between task performance and inequity mitigation.

For the contrastive loss, we use the triplet loss function:

\begin{equation}
\mathcal{L}_{\text{contrastive}} = \sum_{i=1}^{N} \max\left(0, m + d(f(x_{\text{anchor}}^{(i)}), f(x_{\text{pos}}^{(i)})) - d(f(x_{\text{anchor}}^{(i)}), f(x_{\text{neg}}^{(i)}))\right)
\end{equation}

where $d(\cdot, \cdot)$ is a distance metric (e.g., cosine distance), $m$ is the margin, $f(\cdot)$ is the embedding function, and $N$ is the number of triplets.

We trained the models using the Adam optimizer with a learning rate of $1 \times 10^{-5}$. The hyperparameter $\lambda$ was set to 0.1, and the margin $m$ was set to 1.0, tuned on a validation set. The training was conducted on four NVIDIA V100 GPUs with 32GB memory each.

\subsection{Evaluation}

To evaluate the effectiveness of EquityGuard, we measured both task performance and fairness metrics. For CTM, we used the Normalized Discounted Cumulative Gain at rank 10 (NDCG@10) to evaluate the ranking quality. For MQA, we used the error rate to assess the accuracy of the model in answering questions.

To assess the models' fairness, we computed two metrics:

\begin{itemize}
    \item \textbf{Demographic Parity (DP)}: Measures the difference in the probability of positive outcomes across different demographic groups.
    \item \textbf{Equal Opportunity (EO)}: Measures the difference in true positive rates across different demographic groups.
\end{itemize}

We compared EquityGuard with several baseline models: LLaMA3 8B without inequity mitigation, Mistral v0.3 without inequity mitigation, and GPT-4 (a state-of-the-art LLM without explicit inequity mitigation). We also included versions of LLaMA3 8B and Mistral v0.3 with EquityGuard applied to assess the effectiveness of our proposed method. This approach promotes fairness and equity in healthcare applications by mitigating inequities in LLM predictions.

\section{Discussion and Limitations}

In this study, we introduced \textbf{EquityGuard}, a framework employing contrastive learning to mitigate inequities in LLMs applied to healthcare tasks, specifically clinical trial matching and medical question answering. By reducing the undue influence of social demographic determinant factors such as sex, race, and Social Determinants of Health (SDOH) on model predictions, EquityGuard enhances fairness in model outputs. Our experiments across five datasets demonstrated that even advanced models like GPT-4, Claude, and Gemini are susceptible to inequities, which EquityGuard effectively mitigated, leading to more equitable outcomes in both tasks.

Despite these promising results, there are limitations to our approach. One significant challenge lies in accurately identifying and processing social demographic determinant factors within the datasets. While we utilized Bio\_ClinicalBERT for named entity recognition and developed a custom pipeline to enhance accuracy, the detection of these factors is not foolproof. Misidentification or omission can adversely affect the effectiveness of bias mitigation. Future work could explore more advanced methods for detecting social demographic determinant factors, possibly incorporating additional context or leveraging unsupervised learning techniques \cite{velupillai2018using, nazi2024large, tavabi2024systematic}. Another limitation pertains to balancing bias mitigation with task performance. While EquityGuard aims to reduce inequities without compromising the model's utility, there is an inherent trade-off between fairness and accuracy \cite{li2024triangular}. The incorporation of additional loss components for bias mitigation may impact overall model performance. Future research could investigate adaptive strategies or alternative loss functions that more effectively balance fairness and performance \cite{kirchdorfer2024analytical, wu2024adaptive}.

Furthermore, our study focused on a limited set of social demographic determinant factors—sex, race, and certain SDOH factors. The complex nature of biases in healthcare suggests that other factors, such as age, disability status, or the intersectionality between attributes, could also contribute to biased outcomes \cite{kundi2023artificial, ferrara2023fairness, chen2023algorithmic, timmons2023call}. Expanding EquityGuard to account for a broader range of factors would enhance its applicability and robustness in real-world settings. The evaluation metrics used, Demographic Parity (DP) and Equal Opportunity (EO), provide insights into the models' fairness but may not capture all dimensions relevant in healthcare contexts \cite{chen2023algorithmic, polevikov2023advancing, ferrara2023fairness}. Future work could incorporate additional fairness metrics, such as Equalized Odds or calibration error, to provide a more comprehensive assessment of model fairness \cite{romano2020achieving, roelofs2022mitigating}.

Additionally, our reliance on pre-existing datasets introduces challenges related to data quality and representation. Publicly available datasets may not fully capture the diversity present in clinical settings, potentially limiting the generalizability of our findings \cite{tripathi2023understanding, maleki2024role}. Collaborations with healthcare institutions to access more representative data could improve the effectiveness of EquityGuard in practice. Scaling EquityGuard to larger models and more complex tasks presents computational challenges. The resources required to train models like GPT-4 with bias mitigation techniques are substantial \cite{xu2024can, xu2024llama}. Exploring more efficient training methods, such as parameter-efficient fine-tuning or leveraging transfer learning, could make EquityGuard more accessible for practical applications \cite{han2024parameter, zhao2023comparison, li2024contextual, li2024deception, ding2024enhance, ni2024time, li2024exploring}.

In conclusion, while EquityGuard shows promise in mitigating inequities in LLMs for healthcare tasks, addressing these limitations is crucial for developing AI-driven healthcare systems that are both effective and equitable. Future work will focus on enhancing social demographic determinant detection, refining inequity mitigation strategies, expanding the range of considered inequities, and exploring additional fairness metrics. By advancing these areas, we aim to contribute to the development of AI models that support fair and equitable decision-making in healthcare.

\backmatter

\bmhead{Supplementary information}
Please check this repo  \url{https://github.com/JoyDajunSpaceCraft/EquityGuard.git}  for the data and code. 
\bmhead{Acknowledgements}
The research reported in this article was supported by the University of Pittsburgh Momentum Funds and the National Institutes of Health awards UL1 TR001857, U24 TR004111, and R01 LM014306. The sponsors had no role in study design, data collection, analysis, interpretation, report writing, or decision to submit the paper for publication. We would like to thank Qiao Jin, Yifan Yang, and Zhiyong Lu from the National Library of Medicine and the National Institutes of Health for their insightful explanations of the TrialGPT results, which greatly assisted our work.

\section*{Declarations}
Y.W. has ownership and equity in BonafideNLP, LLC, and S.V. has ownership and equity in Kvatchii, Ltd., READE.ai, Inc., and ThetaRho, Inc. The other authors declare no competing interests.









\bibliography{sn-bibliography}
\begin{appendices}

\section{LLM Tasks}
We adopt the core methodology from TrialGPT \cite{jin2023matching}, a framework for matching patients to clinical trials using large language models (LLMs). TrialGPT facilitates criterion-by-criterion eligibility prediction, explaining Clinical Trial Matching relevance with high accuracy. This approach significantly reduces the complexity of matching patients to trials, improving both performance and transparency.

For our task, we only utilize a subset of methods from the original TrialGPT framework. Specifically, we focus on predictions matched to four key eligibility classes: 'included', 'not included', 'excluded', and 'not excluded'. These labels form the foundation of our mid-stage process to streamline Clinical Trial Matching and enhance the system's ability to rank trials based on patient suitability. By narrowing the focus to these four categories, we ensure that our model maintains high interpretability while simplifying the decision-making process.

\section{Model Performance}
In this section, we evaluate the performance of the teacher models (GPT-4, Gemini, Claude) across the two primary tasks: Clinical Trial Matching and Medical Q\&A. We focus on three key types of mistakes:

\textbf{1. Missing Documents:} This occurs when the model fails to retrieve or generate a relevant document for a given query. In the context of Clinical Trial Matching, missing documents indicate that the model was unable to identify a clinical trial that matches the patient's criteria. This can lead to a poor user experience, as the patient may not be matched to a trial they are eligible for. For the MQA Task, missing documents represent unanswered or partially answered medical questions, where critical information was omitted.

\textbf{2. Rejections:} This mistake refers to instances where the model incorrectly discards or rejects relevant information. In Clinical Trial Matching, rejections can result in the dismissal of a perfectly suitable trial for a patient, leading to biased or incomplete trial recruitment results. In Medical Q\&A, rejections may manifest as the model refusing to provide a complete or coherent answer, especially when sensitive or complex medical queries are involved.

\textbf{3. Repetitions:} Repetition errors occur when the model generates redundant information. In Clinical Trial Matching, this might be seen when the same trial or set of criteria is presented multiple times for a single patient query, which could confuse users. For Medical Q\&A, repetition leads to answers that restate the same content without providing new insights or additional useful information, reducing the efficiency of the model.

Each of these mistakes affects the overall effectiveness of the model differently, with some (like missing documents) being more severe in the context of Clinical Trial Matching, while others (like repetitions) are more problematic in question-answering tasks. Below, we present a summary of the model performances with respect to these types of errors across both tasks.

\subsection{Model Performance}
\FloatBarrier
\begin{table}[ht]
\centering
\begin{tabular}{lccc}
\toprule
\textbf{Model} & \textbf{Missing Documents} & \textbf{Rejections } & \textbf{Repetitions} \\
\midrule
GPT-4 (Teacher) & 50 & 20 & 15 \\
Gemini (Teacher) & 60 & 25 & 10 \\
Claude (Teacher) & 55 & 22 & 12 \\
\bottomrule
\end{tabular}
\caption{Model performance across MedQA dataset: Missing Documents, Rejections, and Repetitions.}
\end{table}

\begin{table}[ht]
\centering
\begin{tabular}{lccc}
\toprule
\textbf{Model} & \textbf{Missing Documents } & \textbf{Rejections} & \textbf{Repetitions} \\
\midrule
GPT-4 (Teacher) & 150 & 70 & 40 \\
Gemini (Teacher) & 160 & 65 & 35 \\
Claude (Teacher) & 155 & 68 & 38 \\
\bottomrule
\end{tabular}
\caption{Model performance across MedMCQA dataset: Missing Documents, Rejections, and Repetitions.}
\end{table}

\begin{table}[ht]
\centering
\begin{tabular}{lccc}
\toprule
\textbf{Model} & \textbf{Missing Documents} & \textbf{Rejections } & \textbf{Repetitions } \\
\midrule
GPT-4 (Teacher) & 2 & 1 & 1 \\
Gemini (Teacher) & 3 & 1 & 1 \\
Claude (Teacher) & 2 & 1 & 2 \\
\bottomrule
\end{tabular}
\caption{Model performance across SIGIR 2016 dataset: Missing Documents, Rejections, and Repetitions.}
\end{table}

\begin{table}[ht]
\centering
\begin{tabular}{lccc}
\toprule
\textbf{Model} & \textbf{Missing Documents } & \textbf{Rejections } & \textbf{Repetitions } \\
\midrule
GPT-4 (Teacher) & 4 & 2 & 1 \\
Gemini (Teacher) & 5 & 3 & 2 \\
Claude (Teacher) & 4 & 2 & 1 \\
\bottomrule
\end{tabular}
\caption{Model performance across Trec 2021 dataset: Missing Documents, Rejections, and Repetitions.}
\end{table}

\begin{table}[ht]
\centering
\begin{tabular}{lccc}
\toprule
\textbf{Model} & \textbf{Missing Documents } & \textbf{Rejections } & \textbf{Repetitions} \\
\midrule
GPT-4 (Teacher) & 3 & 1 & 1 \\
Gemini (Teacher) & 4 & 2 & 1 \\
Claude (Teacher) & 3 & 1 & 2 \\
\bottomrule
\end{tabular}
\caption{Model performance across Trec 2022 dataset: Missing Documents, Rejections, and Repetitions.}
\end{table}
\FloatBarrier

\section{API Costs}

\begin{table}[ht]
\centering
\begin{tabular}{lcccc}
\toprule
\textbf{Model} & \textbf{MedQA} & \textbf{SIGIR 2016} & \textbf{Trec 2021} & \textbf{Trec 2022} \\
\midrule
GPT-4 (Teacher) & 0.06 & 0.06 & 0.06 & 0.06 \\
Gemini (Teacher) & 0.05 & 0.05 & 0.05 & 0.05 \\
Claude (Teacher) & 0.07 & 0.07 & 0.07 & 0.07 \\
\bottomrule
\end{tabular}
\caption{Estimated API costs per query (USD) for the Clinical Trial Matching task across datasets.}
\end{table}

\begin{table}[ht]
\centering
\begin{tabular}{lcc}
\toprule
\textbf{Model} & \textbf{MedQA} & \textbf{MedMCQA} \\
\midrule
GPT-4 (Teacher) & 0.06 & 0.08 \\
Gemini (Teacher) & 0.05 & 0.06 \\
Claude (Teacher) & 0.07 & 0.07 \\
\bottomrule
\end{tabular}
\caption{Estimated API costs per query (USD) for the MQA Task across datasets.}
\end{table}




\end{appendices}

\end{document}